\documentclass[letterpaper]{article} 
\usepackage[submission]{aaai25}  
\usepackage{times}  
\usepackage{helvet}  
\usepackage{courier}  
\usepackage[hyphens]{url}  
\usepackage{graphicx} 
\usepackage{natbib}  
\usepackage{amsmath} 
\usepackage{amssymb}
\usepackage{amsthm}
\usepackage{mathtools}
\usepackage{cleveref}
\usepackage{caption} 
\usepackage{algorithm}
\usepackage{algorithmic}
\usepackage{xcolor}
\usepackage{bbding}
\usepackage{pifont}
\usepackage{booktabs}
\usepackage{colortbl}
\usepackage{multicol}
\usepackage{multirow}
\usepackage{subcaption}
\definecolor{my_green}{RGB}{51,102,0}
\definecolor{my_red}{RGB}{204, 0, 0}

\newcommand{\cmark}{\textcolor{my_green}{\ding{51}}} 
\newcommand{\xmark}{\textcolor{my_red}{\ding{55}}} 
\usepackage{newfloat}
\usepackage{listings}
\DeclareCaptionStyle{ruled}{labelfont=normalfont,labelsep=colon,strut=off} 
\floatstyle{ruled}
\newfloat{listing}{tb}{lst}{}
\floatname{listing}{Listing}
\title{CaRDiff: Video Salient Object Ranking Chain of Thought Reasoning\\for Saliency Prediction with Diffusion}
\author{
    Yunlong Tang\textsuperscript{\rm 1,2,}\thanks{Work done during internship at ByteDance.},
    Gen Zhan\textsuperscript{\rm 1},
    Li Yang\textsuperscript{\rm 1},
    Yiting Liao\textsuperscript{\rm 1},
    Chenliang Xu\textsuperscript{\rm 2}
}
\affiliations{

    \textsuperscript{\rm 1}ByteDance\\
    \textsuperscript{\rm 2}University of Rochester\\
}

\usepackage{bibentry}

\begin{document}
\maketitle

\begin{abstract}
Video saliency prediction aims to identify the regions in a video that attract human attention and gaze, driven by bottom-up features from the video and top-down processes like memory and cognition. Among these top-down influences, language plays a crucial role in guiding attention by shaping how visual information is interpreted.
Existing methods primarily focus on modeling perceptual information while neglecting the reasoning process facilitated by language, where ranking cues are crucial outcomes of this process and practical guidance for saliency prediction.
In this paper, we propose CaRDiff (\textbf{Ca}ption, \textbf{R}ank, and generate with \textbf{Diff}usion), a framework that imitates the process by integrating multimodal large language model (MLLM), a grounding module, and a diffusion model, to enhance video saliency prediction. Specifically, we introduce a novel prompting method VSOR-CoT (\textbf{V}ideo \textbf{S}alient \textbf{O}bject \textbf{R}anking \textbf{C}hain \textbf{o}f \textbf{T}hought), which utilizes an MLLM with a grounding module to caption video content and infer salient objects along with their rankings and positions. This process derives ranking maps that can be sufficiently leveraged by the diffusion model to decode the saliency maps for the given video accurately.
Extensive experiments show the effectiveness of VSOR-CoT in improving the performance of video saliency prediction.
The proposed CaRDiff performs better than state-of-the-art models on the MVS dataset and demonstrates cross-dataset capabilities on the DHF1k dataset through zero-shot evaluation.
\end{abstract}
\section{Introduction}
\label{sec:intro}

With the rapid growth of online video platforms, millions of videos are produced and consumed daily. This surge presents new challenges in video processing, such as enhancing video quality to improve user experience and compressing videos to save storage costs. Identifying regions of interest (ROIs) within videos has, therefore, become crucial: on the one hand, quality enhancement can focus on these regions; on the other hand, compression algorithms can be locally applied to non-interest regions. This maximizes users' experience while minimizing video storage and transmission costs.
\begin{figure}[!ht]
    \centering
    \includegraphics[width=\columnwidth]{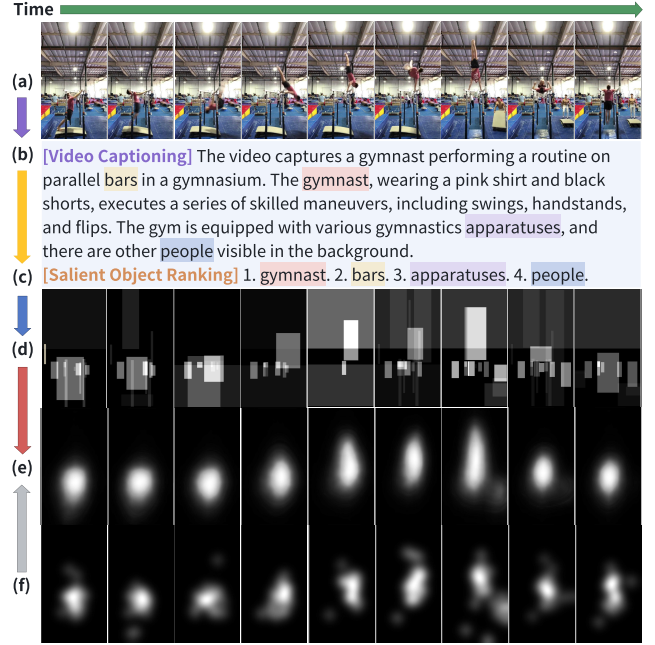}
    \caption{Given (a) the input video, CaRDiff generates (b) video captions and (c) salient objects ranking via VSOR-CoT. These create (d) ranking maps that guide the diffusion model, resulting in (e) saliency predictions, which show accuracy compared to (f) ground-truth saliency maps.}
    \label{fig:teaser}
    \vspace{-1em}
\end{figure}
This growing importance of identifying and prioritizing ROIs in videos has led to increased research on video saliency prediction~\cite{droste2020unified_UNISAL,bellitto2021hierarchical_HD2S,wang2021spatio_STSANet,jain2021vinet,VSFT,xiong2024diffsal,MVFormer}, which aims to predict the regions of a video most likely to capture human attention and gaze.

Though significant progress has been made in this area, as today's video content becomes increasingly rich and scenes become more complex, video saliency prediction models trained on limited-size and limited-scope datasets~\cite{wang2018revisiting_dhf1k,hollywood2} are gradually becoming inadequate. To address this issue, some works have proposed datasets with more diverse scenes, such as the NPF~\cite{yang2023npf} and the MVS dataset~\cite{MVFormer}. However, due to the high costs associated with fixation annotation, the size of these datasets remains limited. Instead of attempting to create a larger dataset, we sought a solution by analyzing the characteristics of the human gaze: human gaze control during real-world scene perception is influenced by bottom-up stimulus-based information and top-down memory-based knowledge from internal visual and cognitive systems~\cite{buswell1935people, yarbus2013eye}.
In most cases, the duration and position of individual fixations are determined by the latter~\cite{loftus1978cognitive, henderson1999effects,henderson2003human}, with language playing a role in this process guiding visual attention.
To involve high-level semantics as an auxiliary, captioning has been leveraged to improve salient object detection~\cite{capsal}. However, they are still limited to the qualities of captions and lack an understanding of the importance of individual objects mentioned in the captions. Additionally, salient object ranking implies that the importance of objects can aid in saliency map prediction~\cite{kalash2019relative,song2023rethinking}. Therefore, we believe the role of language in human gaze control is primarily to reason out a salient object ranking, which assists the visual system in locating salient regions.

Based on the above analysis, we propose CaRDiff (\textbf{Ca}ption, \textbf{R}ank, and generate with \textbf{Diff}usion), a framework designed to enhance video saliency prediction by integrating a multimodal large language model (MLLM) to capture high-level semantics explicitly. Specifically, we introduce a novel prompting method named VSOR-CoT (\textbf{V}ideo \textbf{S}alient \textbf{O}bject \textbf{R}anking \textbf{C}hain \textbf{o}f \textbf{T}hought). Instead of only generating captions as an auxiliary~\cite{capsal}, VSOR-CoT leverages the strong reasoning capability of MLLM to explicitly derive salient object ranking through chain-of-thought reasoning after generating high-quality video captions. Unlike the implicit way~\cite{capsal} to convey the semantic cues with an attention map, we introduce ranking maps, as shown in \Cref{fig:teaser} (d) to represent locations and ranking cues of salient objects with bounding boxes and grayscale values, where the locations are captured by a grounding module and the ranking are derived from VSOR-CoT. Combined with the video frames, these ranking maps serve as decoding conditions for a diffusion model to predict the final saliency maps. The proposed CaRDiff achieves state-of-the-art performance on MVS~\cite{MVFormer}. It also shows cross-dataset capability on the popular video saliency prediction benchmark DHF1k through zero-shot evaluation.

In short, our main contributions are as follows:

\begin{itemize}
\item We propose an s innovative video saliency prediction framework, CaRDiff. Based on the powerful reasoning capacity of MLLM, salient objects with their ranking can be reasoned out after video captioning through the proposed VSOR-CoT.
\item We introduce ranking maps to maintain the position and ranking cues of salient objects, which are derived from reasoning results of VSOR-CoT and seamlessly guide the diffusion process to enhance saliency prediction.
\item Experimental results show that the proposed CaRDiff achieves state-of-the-art performance on the MVS dataset and cross-dataset capability on the popular video saliency prediction benchmark DHF1k through zero-shot evaluation.
\end{itemize}

\section{Related Work}
\label{sec:related_work}
\subsubsection{Video Saliency Prediction and Ranking.}
Saliency detection aims to predict the regions in an image or video that attract human attention. Early works about video saliency prediction include ITTI~\cite{itti} and GBVS~\cite{GBVS}. In recent years, deep learning-based methods have significantly progressed in video saliency prediction~\cite{salicon,pan2017salgan, Jiang2020DeepVS20AS, stranet, droste2020unified_UNISAL, wang2021spatio_STSANet, MVFormer}.
Some methods~\cite{liu2020learning_va, tsiami2020stavis, yang2023npf, xiong2024diffsal} utilize multimodal information in videos to improve the performance of video saliency prediction. These methods mainly focus on fusing the visual and audio signals to predict the saliency map.
Saliency Object Ranking aims to rank the salient objects in an image or video according to their saliency levels~\cite{islam2018revisiting}. It utilizes more semantics from videos and has potential benefits for video saliency prediction tasks~\cite{kalash2019relative,song2023rethinking}.

\subsubsection{Multimodal LLM and Chain-of-Thought.}
Multimodal Large Language Models (MLLMs) have achieved significant progress in various multimodal tasks. Some methods~\cite{zhang2023videollama,liu2024llava,wang2023caption,tang2024avicuna,hua2024v2xum} use MLLMs to generate captions and summmarizations for images or videos. Others~\cite{chen2023shikra,peng2023kosmos,xuan2024pink,hua2024finematch} use MLLMs to infer the relationships between objects in images. They have also extended to the video domain for video understanding~\cite{tang2023video}.
The chain of thought is a prompting technique for LLMs/MLLMs that breaks down the reasoning process into multiple smaller steps, improving the accuracy of the results~\cite{wei2022chain}. For instance, when an LLM is asked to solve a math problem, it might make a mistake if asked for the answer directly. However, if the LLM generates a step-by-step solution process before arriving at the final answer, the accuracy increases.

\subsubsection{Diffusion Model.}
Recent advancements in diffusion models have significantly bolstered their efficacy in generative modeling~\cite{ramesh2022hierarchical,Song_2023_CVPR}. Denoising diffusion probabilistic models (DDPM)~\cite{ho2020denoising_ddpm} and denoising diffusion implicit models~\cite{song2020denoising_ddim} have provided a robust framework for iterative noise addition and removal. Architectural enhancements, notably incorporating attention mechanisms inspired by Transformer models, have further improved model performance~\cite{peebles2023dit}. Apart from generative tasks, diffusion models can also be used for discriminative tasks, e.g., object detection~\cite{chen2023diffusiondet} and segmentation~\cite{amit2021segdiff}. Moreover, text-image alignment has been utilized to enhance the performance of diffusion models in various computer vision tasks, like detection and segmentation~\cite{kondapaneni2024text}.
Diff-Sal~\cite{xiong2024diffsal} is a pioneer work that adopts a diffusion model to generate saliency maps conditioned by input video and audio, showing promising results compared to conventional methods.

\section{Method: CaRDiff}
\label{sec:method}

In this section, we introduce our data curation method for VSOR-CoT Tuning data construction and CaRDiff's architecture and training strategies.

\subsection{Data Curation}
\begin{figure}
    \centering
    \includegraphics[width=\columnwidth]{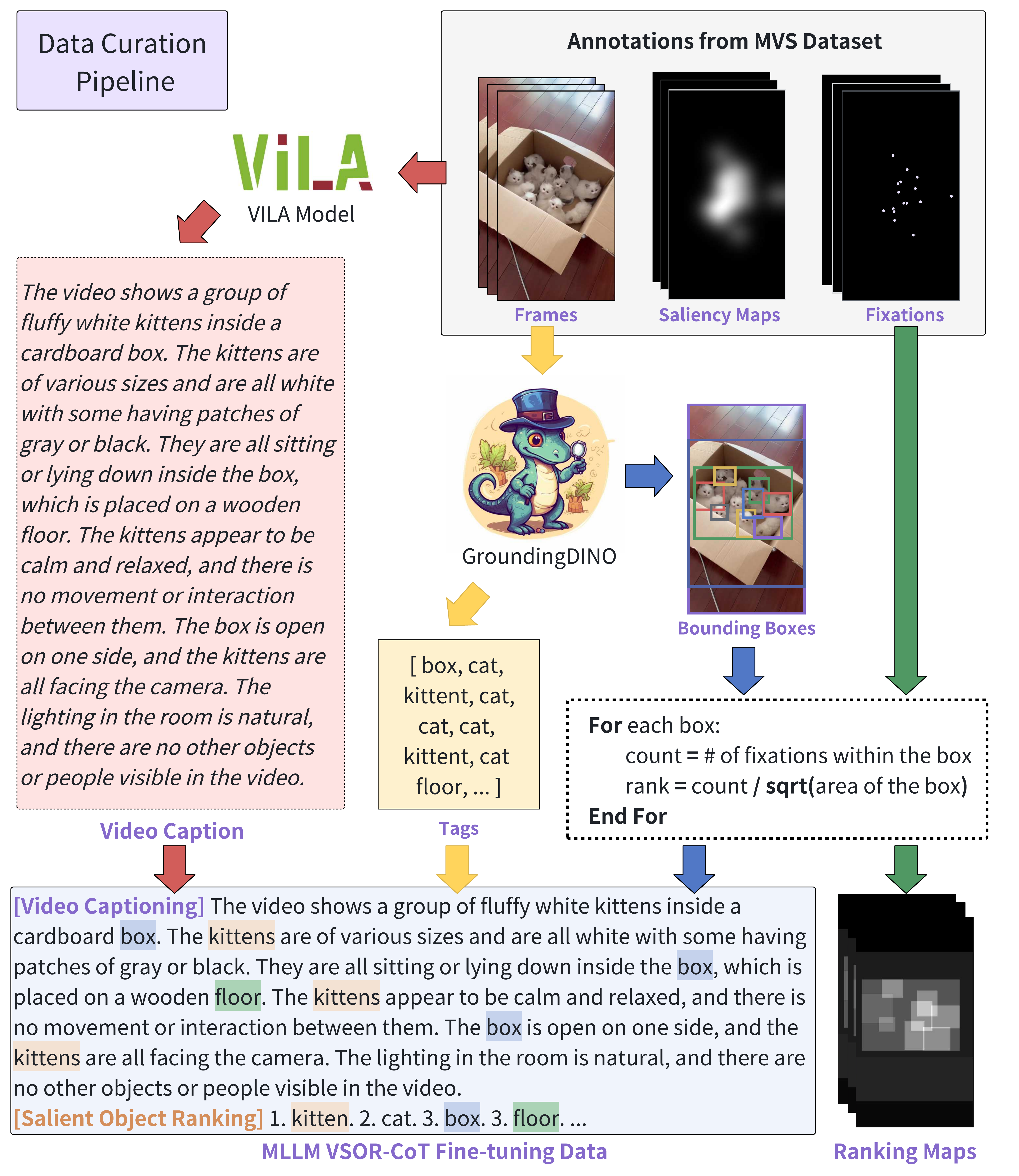}
    \caption{The pipeline of data curation.}
    \label{fig:data}
    \vspace{-1.5em}
\end{figure}

Our data curation pipeline is shown in \Cref{fig:data}.
It is mainly based on the MVS~\cite{MVFormer}, a dataset with videos across abundant scenarios, which contains 1007 video clips, each with fixation maps and saliency maps annotations. We first use Recognize Anything~\cite{zhang2024recognize} and GroundingDINO~\cite{liu2023groundingdino} to extract objects from each video frame, including the objects' position represented by bounding box coordinates and tags. The objects' tags $O=\{o_1,o_2,...,o_n\}$ and positions $B=\{b_1,b_2,...,b_n\}$ are combined with the number of gaze points in the fixation map inside the bounding box to calculate the salient ranking of the object $r_i$, which is computed as:
\begin{equation}
r_i=r(b_i, M^f) = \frac{1}{\sqrt{|b_i|}} \sum_{(u,v) \in b_i} \mathbb{I}[M^f(u,v)>0],
\end{equation}
where $ b_i $ denotes the set of pixels within the bounding box of the object $ i $; $M^f$ is the fixation map; $ |b_i| $ measures the spatial size of $ b_i $, calculated as width $\times$ height of the bounding box; $ \mathbb{I}[M^f(u,v)>0] $ indicates whether there is a fixation point at the spatial coordinate $ (u,v) $. The higher the value of $r_i$, the higher the object's salient. We then use this object information to generate the ranking response, which is part of the response output by the MLLM, i.e., the complement of VSOR-CoT. The ground-truth responses include two parts: (1) video captions and (2) salient object rankings. The latter is from the process above, while the former is generated by VILA-1.5~\cite{lin2024vila}, the state-of-the-art open-source video-language model.

The rankings and positions determine the grayscale value of each pixel in the ranking maps, as defined by the equation:
\begin{equation}
M^r(u,v) = \sum_{i} r_i \cdot \mathbb{I}[(u,v) \in b_i].
\end{equation}
The resulting values are then scaled to the range \([0, 255]\). The ranking map \(M^r\) will be utilized during training, and it serves as a benchmark for comparison with the predicted ranking map for analytical purposes. Unlike relying on masks to represent the locations of salient objects in SOR, in the ranking maps, we only need the bounding boxes of the objects.

\begin{figure*}
    \centering
    \includegraphics[width=\textwidth]{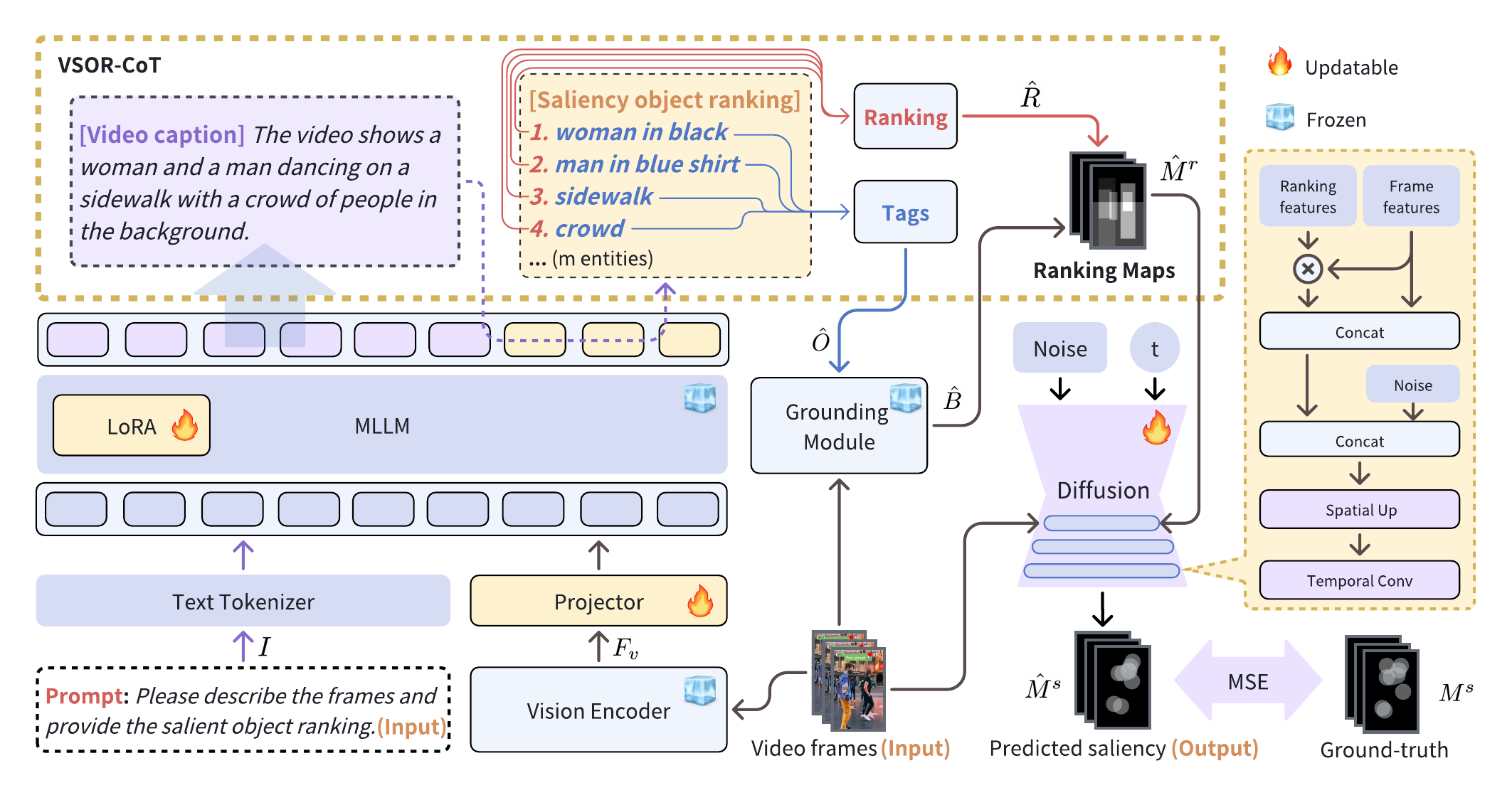}
    \caption{The proposed CaRDiff consists of an MLLM with VSOR-CoT, a grounding module, and a diffusion model.}
    \label{fig:model}
    \vspace{-1.5em}
\end{figure*}

\subsection{Architecture}
\subsubsection{Overview.}
The proposed CaRDiff (\textbf{Ca}ption, \textbf{R}ank, and generate with \textbf{Diff}usion), as shown in \Cref{fig:model}, which consists of an MLLM, a grounding module, and a diffusion model. The MLLM generates the caption for the video content and drives salient object ranking. The grounding module captures the locations of the salient objects. The ranking and the locations are then used to generate ranking maps. The diffusion takes the ranking maps and the video frames as conditions to decode the corresponding saliency maps.

\subsubsection{MLLM with VSOR-CoT.}
Our MLLM is built upon LLaVA~\cite{liu2024llava}. For an input video consisting of $L$ frames, the visual features $F_v=\{f_1,f_2,...,f_L\}$ are first extracted using a pre-trained vision encoder, specifically CLIP ViT-14/L~\cite{radford2021learning_clip}. These features are then processed through a Projector, a linear layer designed to align the visual semantic space with the LLM's input space. A text instruction $I$ is tokenized and provided to the LLM along with the visual features. The LLM then generates a video caption $C$ and a salient object ranking list $S$ based on the inputs $F_v$ and $I$:
\begin{equation}
(C,S) = LLM(I,F_v).
\end{equation}
The salient object ranking list $S=\{(\hat{o}_i,\hat{r}i)\}_{i=1}^{m}$ is derived from the caption $C$ through the VSOR-CoT process during LLM inference. This ranking list $S$ includes a set of salient object tags $\hat{O}=\{\hat{o}_1,\hat{o}_2,...,\hat{o}_m\}$ and their corresponding rankings $\hat{R}=\{\hat{r}_1,\hat{r}_2,...,\hat{r}_m\}$, as illustrated in \Cref{fig:model}. Specifically, during LLM inference, the probability of the ranking list $S$ can be decomposed as follows:
\begin{equation}
\label{eq:vsor-cot}
\begin{aligned}
p(S|I, F_v) &= \prod_{i=1}^{m} p(s_i|s_{<i}, I, F_v) \\
&= \prod_{i=1}^{m} \sum_{C} p(s_i|s_{<i}, C, I, F_v) \cdot p(C|s_{<i}, I, F_v),
\end{aligned}
\end{equation}
where the probability of each rank $s_i$ is calculated based on the preceding ranks $s_{<i}$, the caption $C$, the instruction $I$, and the visual features $F_v$. The VSOR-CoT method ensures that the ranking list $S$ is generated in a contextually coherent manner by considering the interactions between the video caption and the salient objects.

\subsubsection{Grounding Module.}
Our grounding module $G$ is based on GroundingDINO~\cite{liu2023groundingdino} to capture the locations of the salient objects predicted by VSOR-CoT. It can take textual tags as prompts to ground objects in images or frames, where the prompts are from the tags $\hat{O}$ predicted by MLLM:
\begin{equation}
    \hat{B}=G(\hat{O},F),
\end{equation}
 where $ \hat{B}=\{\hat{b}_1, \hat{b}_2, ..., \hat{b}_m\}$ is the set of locations (bounding boxes) of these salient objects.

\subsubsection{Ranking Map Synthesis.}
The predicted ranking map $ \hat{M}^r $ is synthesized according to the predicted objects' locations $\hat{B}$ and their ranking $\hat{R}$:
\begin{equation}
\hat{M}^r(u,v) = \sum_{i} \hat{r}_i^* \cdot \mathbb{I}[(u,v) \in \hat{b}_i],~~\hat{r}^*_i=1-\frac{\hat{r_i}-1}{m-1},
\end{equation}
which is a bit different from the ranking map $M^r$ in annotations.
The ranking map $ \hat{M}^r $ will have regions corresponding to these objects with their respective grayscale intensities.
\begin{figure*}[!ht]
    \centering 
    \includegraphics[width=\linewidth]{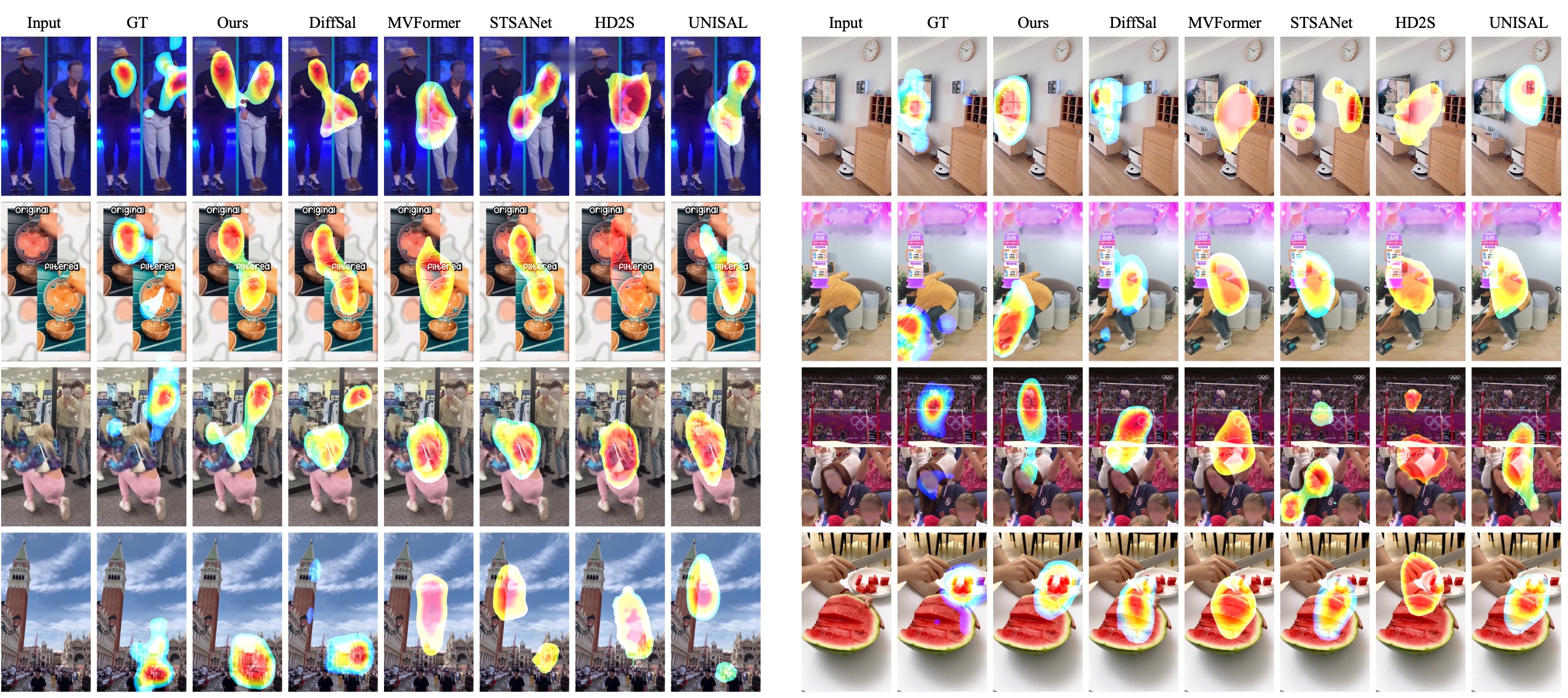}
    \caption{Results Visualization. Our CaRDiff shows advantages across multiple state-of-the-art models, especially in videos with rich content and complex scenarios. More results visualized can be found in the Appendix.}
    \label{fig:compare}
    \vspace{-1.5em}
\end{figure*}
\subsubsection{Saliency Diffusion Prediction.}
The diffusion model in CaRDiff involves two main phases: the forward and reverse denoising processes. In the forward process, noise is added to the saliency map $ M^s $ over a series of time steps $ t $:
\begin{equation}
q(M^s_t | M^s_0) = \mathcal{N}(M^s_t; \sqrt{\alpha_t} M^s_0, (1 - \alpha_t) \textbf{I}),
\end{equation}
where $ M_0 = M^s $. At each time step, noise $ \epsilon \sim \mathcal{N}(0, \textbf{I}) $ is added:
\begin{equation}
M^s_t = \sqrt{\alpha_t} M^s_0 + \sqrt{1 - \alpha_t} \epsilon.
\end{equation}
In the reverse process, the goal is to remove the noise using a U-Net-based diffusion model $D_{\theta}$ that predicts the de-noised saliency map $\hat{M}^s_{t-1}$:
\begin{equation}
\hat{M}^s_{t-1} = \frac{M^s_t}{\sqrt{\alpha_t}} - \frac{1 - \alpha_t}{\sqrt{(1 - \alpha_t)\alpha_t}} D_{\theta}(M^s_t, t, \hat{M}^r\otimes F_{v'}),
\end{equation}
where $F_{v'}$ is video features, $\hat{M}^r$ is predicted ranking map, and $\otimes$ is the position-wise product and concatenation.

\subsection{Three-Stages Training}
CaRDiff's training process includes three stages: modality alignment, CoT tuning, and diffusion training.

\subsubsection{Modality Alignment.}
Modality alignment involves aligning visual features with the input space of the LLM (Language Model). During this stage, only a single-layer projector is updated throughout the training process. The training task is image captioning, utilizing the LCS-558K dataset~\cite{liu2024llava}, which is a subset comprising 558K image-text pairs from LAION-CC-SBU with BLIP-generated captions~\cite{li2023blip}. The instruction prompts the MLLM to generate captions for given images. The loss function employed in this process is the cross-entropy loss, defined as follows:
\begin{equation}
\label{eq:ce}
\mathcal{L}_{CE} = -\sum_{t=1}^{T} \log P(y_t | y_{<t}, I),
\end{equation}
where $ y_t $ represents the t-th word in the caption, and $ I $ denotes the visual features extracted from the image. This cross-entropy loss encourages the model to generate captions that closely match the ground truth descriptions.
\subsubsection{CoT Tuning.}
At this stage, multiple video frame features will replace the input image feature in the first stage. The instruction prompts the MLLM to predict video captions and the saliency object ranking following the \Cref{eq:vsor-cot}. The parameters of the Projector, vision encoder, and MLLM are frozen. The fine-tuning is supervised by the annotations obtained from the data curation process. The loss function also follows \Cref{eq:ce}.

\subsubsection{Diffusion Training.}
The diffusion model is trained by minimizing the Mean Squared Error (MSE) loss between the predicted noise and the actual noise added during the forward diffusion process:
\begin{equation}
\mathcal{L}_{\text{MSE}} = \mathbb{E}_{M_0, \epsilon, t} \left[ \left\| M_0 - D_{\theta}(M^s_t, t, \hat{M}^r, F_{v'}) \right\|_2^2 \right],
\end{equation}
where $ M_0 = M^s $ and $ \epsilon $ is the actual noise. This loss function ensures the neural network accurately predicts the noise, allowing effective denoising in the reverse process.

By encoding the ranks of salient objects in a grayscale ranking map and integrating these with the video frames, the CaRDiff framework guides the diffusion model to focus on regions of higher importance, resulting in an accurate and semantically meaningful saliency map.
\section{Experiments}
In this section, we introduce the experimental evaluations of our CaRDiff method against state-of-the-art models. Our experiments include the comparison of performance on the MVS dataset, ablation study, ranking map ratio experiments, ranking correlation analysis, ranking map replacement experiments, and cross-dataset zero-shot evaluation.
\subsection{Experiment Setups}
\subsubsection{Datasets.} We evaluate our method on the MVS dataset with rich video contents and complex scenarios for the maintenance comparison. We also evaluate our method on the validation set of DHF1k~\cite{wang2018revisiting_dhf1k} datasets for cross-dataset zero-shot evaluation. 

\begin{table*}[!ht]
\centering
\resizebox{\textwidth}{!}{%
\begin{tabular}{l|cccc|cccc}
\toprule
\multirow{2}{*}{\textbf{\begin{tabular}[c]{@{}l@{}}~\\ Methods\end{tabular}}} & \multicolumn{4}{c|}{\textbf{Attributes}} & \multicolumn{4}{c}{\textbf{Performance}} \\ \cline{2-9} 
 & \begin{tabular}[c]{@{}c@{}}Video-\\ based\end{tabular} & \begin{tabular}[c]{@{}c@{}}Re-\\ trained\end{tabular} & Modality & \begin{tabular}[c]{@{}c@{}}Loss\\ function\end{tabular} & AUC-J & CC & Sim & NSS \\ \midrule\midrule
ITTI \cite{itti} & \xmark & \xmark & V & Non-DL & 0.783 & 0.435 & 0.464 & 0.978 \\
GBVS \cite{GBVS} & \xmark & \xmark & V & Non-DL & 0.808 & 0.492 & 0.491 & 1.097 \\
SALICON \cite{salicon} & \xmark & \cmark & V & KLD, NSS, Sim & 0.814 & 0.523 & 0.512 & 1.261 \\
AWS-D \cite{aws-d} & \cmark & \xmark & V & Non-DL & 0.675 & 0.240 & 0.384 & 0.560 \\
SalGAN \cite{pan2017salgan} & \xmark & \cmark & V & MSE, BCE & 0.812 & 0.511 & 0.503 & 1.269 \\
SAM \cite{sam} & \xmark & \cmark & V & CC, NSS, KLD & 0.818 & 0.531 & 0.522 & 1.274 \\
DeepVS \cite{Jiang2020DeepVS20AS} & \cmark & \xmark & V & KLD & 0.811 & 0.475 & 0.496 & 1.160 \\
ACLNet \cite{ACLNet} & \cmark & \cmark & V & CC, NSS, KLD & 0.821 & 0.542 & 0.524 & 1.251 \\
STRA-Net \cite{stranet} & \cmark & \cmark & V & KLD, NSS, Sim, CC & 0.826 & 0.563 & 0.531 & 1.289 \\
SalEMA \cite{linardos2019simple_salema} & \cmark & \cmark & V & BCE & 0.835 & 0.591 & 0.544 & 1.326 \\
TASED \cite{min2019tased} & \cmark & \cmark & V & KLD & 0.850 & 0.638 & 0.576 & 1.486 \\
ESAN \cite{chen2021video_esan} & \cmark & \cmark & V & KLD, NSS, Sim, CC & 0.853 & 0.645 & 0.590 & 1.517 \\
UNISAL \cite{droste2020unified_UNISAL} & \cmark & \cmark & V & CC, NSS, KLD & 0.855 & 0.654 & 0.586 & 1.524 \\
HD2S \cite{bellitto2021hierarchical_HD2S} & \cmark & \cmark & V & KLD & 0.858 & 0.662 & 0.603 & 1.550 \\
STSANet \cite{wang2021spatio_STSANet} & \cmark & \cmark & V & KLD, CC & 0.856 & 0.657 & 0.594 & 1.555 \\
ViNet \cite{jain2021vinet} & \cmark & \cmark & V & KLD & 0.857 & 0.664 & 0.595 & 1.561 \\
VSFT \cite{VSFT} & \cmark & \cmark & V & KLD, NSS, Sim, CC & 0.857 & 0.666 & 0.597 & 1.572 \\
Diff-Sal~\cite{xiong2024diffsal} & \cmark & \cmark & V, A & MSE & 0.852 & 0.626 & 0.577 & 1.591 \\
MVFormer~\cite{MVFormer} & \cmark & \cmark & V & KLD, NSS, Sim & 0.864 & 0.687 & 0.614 & 1.646 \\
\rowcolor[HTML]{DAE8FC}CaRDiff (ours) & \cmark & \cmark & V, L & CE, MSE & \textbf{0.870} & \textbf{0.714} & \textbf{0.630} & \textbf{1.685} \\ \bottomrule
\end{tabular}%
}
\caption{Performance comparison of CaRDiff with various state-of-the-art methods on the MVS dataset, demonstrating superior results across key metrics such as AUC-J, CC, SIM, and NSS. \textbf{V}, \textbf{A}, and \textbf{L} indicate vision, audio, and language, respectively.}
\label{tab:comparison}
\vspace{-1.5em}
\end{table*}

\subsubsection{Evaluation Metrics.}
We use several standard metrics to evaluate the performance of saliency prediction models:
Correlation Coefficient (CC) measures the linear relationship between the predicted and ground truth saliency maps, indicating how well the model's predictions match the actual data.
Normalized Scanpath Saliency (NSS) evaluates the predicted saliency map using human fixation data, assessing how well the model predicts where humans are likely to focus.
Similarity Metric (SIM) measures the similarity between the predicted and ground truth saliency maps.
Area Under the Curve-Judd (AUC-J) assesses the model's ability to predict human fixation points, using a metric that combines the true positive rate and false positive rate to evaluate performance.
The detailed calculation methods can be found in the Appendix.

\subsubsection{Baseline Methods.}
We compare our method with various state-of-the-art approaches, including ITTI \cite{itti}, GBVS \cite{GBVS}, SALICON \cite{salicon}, AWS-D \cite{aws-d}, SalGAN \cite{pan2017salgan}, SAM \cite{sam}, DeepVS \cite{Jiang2020DeepVS20AS}, ACLNet \cite{ACLNet}, STRA-Net \cite{stranet}, SalEMA \cite{linardos2019simple_salema}, TASED \cite{min2019tased}, ESAN \cite{chen2021video_esan}, UNISAL \cite{droste2020unified_UNISAL}, HD2S \cite{bellitto2021hierarchical_HD2S}, STSANet \cite{wang2021spatio_STSANet}, ViNet \cite{jain2021vinet}, VSFT \cite{VSFT}, Diff-Sal~\cite{xiong2024diffsal}, and MVFormer~\cite{MVFormer}.

\subsubsection{Implementation Details.}
The LLM adopted is Vicuna-v1.5-7B~\cite{touvron2023llama2}. At the modality alignment, we fine-tune the projector layer for two epochs with a learning rate of $1\times 10^{-3}$. At the CoT tuning stage, we fine-tune the LoRA in the LLM with a learning rate of $1\times 10^{-4}$ for two epochs. Both the first stage and the second stage are conducted on one NVIDIA A100 80G GPU. At the subsequent diffusion training stage, we use four NVIDIA V100 32G GPUs for three epochs, with a learning rate of $1\times 10^{-4}$. The encoder for frames and ranking maps is MViT-v2~\cite{li2022mvitv2} pre-trained on the K400 dataset~\cite{kay2017kinetics400}. The decoded saliency maps are $384\times 224$ for the MVS dataset.

\subsection{Experimental Results} 

\subsubsection{Performance Comparison.} 

The performances of baseline models and our CaRDiff are shown in \Cref{tab:comparison}. Our method achieves state-of-the-art performance across all evaluation metrics on the MVS dataset. Specifically, CaRDiff achieves the highest CC, NSS, SIM, and AUC-J scores, demonstrating its superior ability to predict saliency in mobile videos.

\begin{figure*}[!ht]
    \centering
    \includegraphics[width=\linewidth]{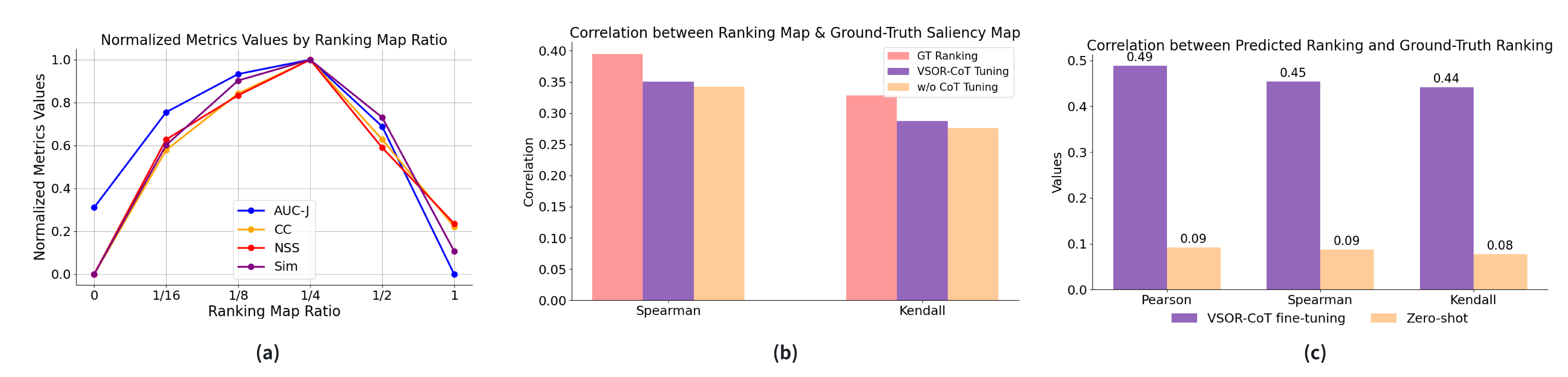}
    \vspace{-1em}
    \caption{(a) Ranking Map Ratio Experiments. (b) and (c): analysis of the ranking-saliency correlation in CaRDiff, illustrating the high correlation between the predicted ranking maps and saliency maps. (b) and (c) show map correlation and rank correlation, respectively.}
    \label{fig:ratio}
    \vspace{-1.5em}
\end{figure*}

\subsubsection{Cross-dataset Performance.}
To evaluate the generalizability of our model, we test its zero-shot performance on unseen datasets. Our model maintains performance without any fine-tuning on DHF1k, indicating its robustness and adaptability to different types of video content.

\begin{table}[!ht]
\centering
\resizebox{0.9\columnwidth}{!}{%
\begin{tabular}{l|cccc}
\toprule
\textbf{Model} & \textbf{AUC-J} & \textbf{CC} & \textbf{Sim} & \textbf{NSS} \\ \midrule\midrule
Diff-Sal & 0.802 & 0.218 & 0.192 & 1.069 \\
MVFormer & 0.844 & 0.299 & 0.198 & 1.501 \\
\rowcolor[HTML]{DAE8FC}CaRDiff (ours) & \textbf{0.845} & \textbf{0.312} & \textbf{0.235} & \textbf{1.584} \\ \bottomrule
\end{tabular}%
}
\caption{Cross-dataset zero-shot evaluation. The models are trained on the MVS dataset and tested on the DHF1k dataset.}
\label{tab:cross}
\vspace{-1em}
\end{table}

\subsubsection{Ablation Study.}

We conduct an ablation study to assess the contribution of each component in our proposed method, as shown in \Cref{tab:ablation}. The results indicate that VSOR-CoT significantly enhances performance, regardless of whether VSOR-CoT fine-tuning (FT) or zero-shot CoT (ZS) is employed. Notably, FT outperforms ZS, underscoring the necessity of CoT Tuning. Additionally, although the ranking maps play a crucial transitional role and cannot be directly ablated, they can be replaced. Thus, we designed the following Ranking Map Replacement Experiments.

\begin{table}[!ht]
\centering
\resizebox{0.9\columnwidth}{!}{%
\begin{tabular}{l|cccc}
\toprule
\textbf{Setting} & \textbf{AUC-J} & \textbf{CC} & \textbf{NSS} & \textbf{Sim} \\ \midrule\midrule
\rowcolor[HTML]{ECF4FF} 
FT w/ VSOR-CoT & \textbf{0.870} & \textbf{0.714} & \textbf{1.685} & \textbf{0.630} \\
FT w/o VSOR-CoT & 0.864 & 0.700 & 1.614 & 0.624 \\
ZS w/ VSOR-CoT & 0.855 & 0.659 & 1.515 & 0.590 \\
ZS w/o VSOR-CoT & 0.846 & 0.626 & 1.459 & 0.577 \\ \bottomrule
\end{tabular}%
}
\caption{Results of the ablation study for CaRDiff, highlighting the impact of different components on performance metrics AUC-J, CC, NSS, and Sim.}
\label{tab:ablation}
\vspace{-1.2em}
\end{table}

\subsubsection{Ranking Map Ratio Experiments.}
We noticed that the different ratios of ranking maps used influenced the results. Therefore, we test different ranking map ratios of 0, 1/16, 1/8, 1/4, 1/2, and 1 on the MVS dataset. The results are shown in \Cref{fig:ratio} (a) and \Cref{tab:ratio}, showing applying ranking maps to 1/4 input frames achieves the highest performance.

\begin{table}[!ht]
\centering
\resizebox{0.90\columnwidth}{!}{%
\begin{tabular}{c|cccc}
\toprule
\textbf{Ranking Map Ratio} & \textbf{AUC-J} & \textbf{CC} & \textbf{NSS} & \textbf{Sim} \\ \midrule\midrule
0 & 0.867 & 0.700 & 1.645 & 0.621 \\
1/16 & 0.869 & 0.709 & 1.670 & 0.626 \\
1/8 & \textbf{0.870} & 0.712 & 1.678 & 0.629 \\
\rowcolor[HTML]{DAE8FC} 
1/4 & \textbf{0.870} & \textbf{0.714} & \textbf{1.685} & \textbf{0.630} \\
1/2 & 0.869 & 0.710 & 1.668 & 0.628 \\
1 & 0.866 & 0.704 & 1.654 & 0.622 \\ \bottomrule
\end{tabular}%
}
\caption{Performance analysis of CaRDiff at different ranking map ratios, showcasing the method's effectiveness in various settings on the MVS dataset.}
\label{tab:ratio}
\vspace{-1em}
\end{table}

\subsubsection{Ranking Map Replacement Experiments.}
\label{sec:replace}
\begin{figure}[!ht]
    \centering
    \includegraphics[width=\columnwidth]{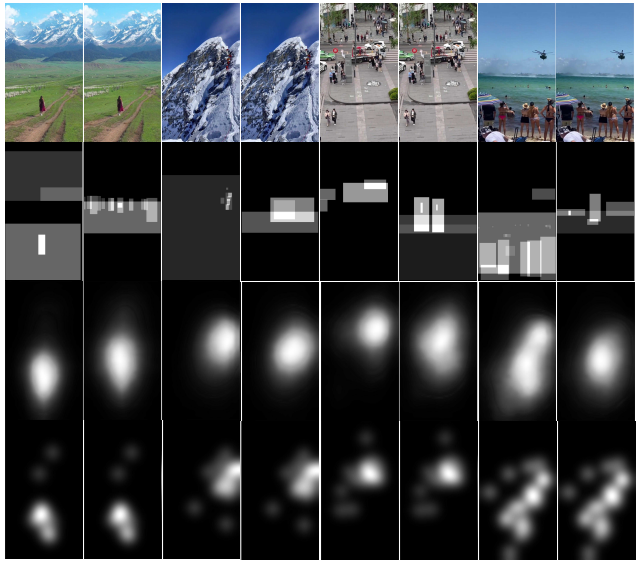}
    \caption{Ranking Map Replacement Experiments. The left columns are the results of adopting a ranking map predicted by MLLM, while the right columns' are replaced by random ranking maps. The latter shows accuracy decreasing, indicating the guidance function of ranking maps.}
    \label{fig:replace}
    \vspace{-1.5em}
\end{figure}

We also conduct ranking map replacement experiments by replacing the predicted ranking maps with random ranking maps, which causes saliency shifts as shown in \Cref{fig:replace}. This further proves the guidance of accurate ranking maps is crucial.

\subsubsection{Ranking-Saliency Correlation Analysis}
We perform a ranking-saliency correlation analysis to understand how well the predicted saliency maps correlate with human eye-tracking data (also shown in (b) and (c) of \Cref{fig:ratio}). The high correlation indicates that our model accurately captures regions of interest, demonstrating the rankings or ranking maps predicted by FT are more closely aligned with the ground truth compared to those generated by ZS.
    

    

\begin{figure}[!ht]
    \centering
    \includegraphics[width=\columnwidth]{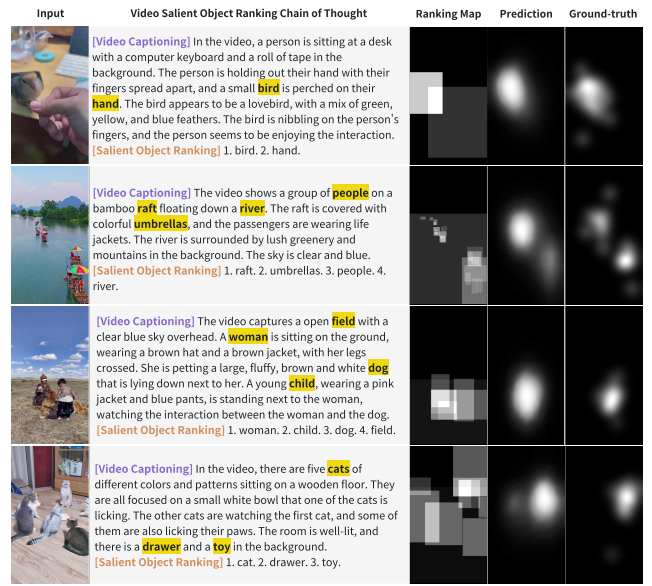}
    \caption{More Visualization Results.}
    \label{fig:exp2}
\end{figure}

\section{Conclusion}
In this paper, we introduced CaRDiff, a novel framework for video saliency prediction that leverages the reasoning capabilities of multimodal large language models. By integrating video captioning and the innovative VSOR-CoT method, CaRDiff effectively ranks salient objects and generates ranking maps that guide diffusion models to predict accurate saliency maps. Our approach outperforms existing state-of-the-art methods on the MVS dataset and demonstrates strong cross-dataset generalization on the DHF1k benchmark. These results validate the effectiveness of incorporating high-level semantics and reasoning in enhancing video saliency prediction.

\bibliography{aaai25}
\appendix
\section{Appendix}

\subsection{Detailed Metrics}
We evaluate the performance of saliency prediction models using several standard metrics~\cite{islam2018revisiting}: Correlation Coefficient (CC), Normalized Scanpath Saliency (NSS), Similarity Metric (SIM), and Area Under the Curve-Judd (AUC-J).

\subsubsection{Correlation Coefficient (CC)} measures the linear relationship between the predicted ($S_p$) and ground truth ($S_g$) saliency maps:
\begin{equation}
CC = \frac{\sum (S_p - \bar{S_p})(S_g - \bar{S_g})}{\sqrt{\sum (S_p - \bar{S_p})^2 \sum (S_g - \bar{S_g})^2}},
\end{equation}
where $\bar{S_p}$ and $\bar{S_g}$ are the mean values of the predicted and ground truth saliency maps, respectively. CC indicates how well the model captures the general spatial distribution of saliency.

\subsubsection{Normalized Scanpath Saliency (NSS)} evaluates $S_p$ using human fixation data ($F$):
\begin{equation}
NSS = \frac{1}{N} \sum \frac{S_p - \mu_{S_p}}{\sigma_{S_p}} F ,
\end{equation}
where $N$ is the number of fixation points, $\mu_{S_p}$ is the mean, and $\sigma_{S_p}$ is the standard deviation of the predicted saliency map. Considering prediction variability, NSS measures the alignment between the predicted saliency map and human fixations.

\subsubsection{Similarity Metric (SIM)} measures the similarity between $S_p$ and $S_g$:
\begin{equation}
SIM = \sum \min(S_p, S_g).
\end{equation}
Values range from 0 to 1, with 1 indicating perfect similarity. SIM compares the saliency distribution at each pixel, providing a measure of local similarity between predicted and ground truth saliency maps.

\subsubsection{Area Under the Curve-Judd (AUC-J)} assesses the model's ability to predict human fixation points:
\begin{equation}
AUC\text{-}J = \int_{0}^{1} TPR(FPR) \, d(FPR), 
\end{equation}
where $TPR$ is the true positive rate, and $FPR$ is the false positive rate. AUC-J evaluates the model's effectiveness in distinguishing between true human fixation points and random points, assessing how well predicted saliency regions correspond to actual human fixations.


\subsection{More Visualization Results}
\begin{figure}[!ht]
    \centering
    \includegraphics[width=\columnwidth]{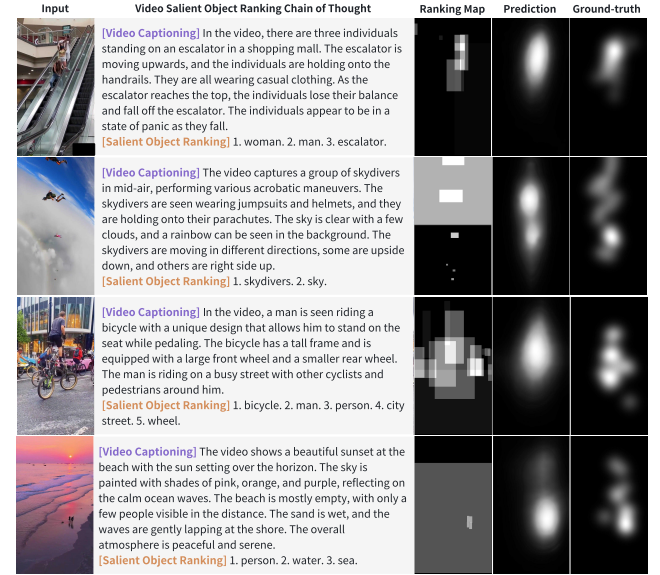}
    \caption{More Visualization Results.}
    \label{fig:exp4}
\end{figure}

We visualize more results predicted by CaRDiff in \Cref{fig:exp4} to the model's advantage in using language to prioritize and accurately predict attention-worthy regions in videos.

\subsection{Limitations and Future Work}
The performance relies on the accuracy of the grounding module, and the combination of an MLLM and a diffusion model increases computational complexity, making it less feasible for real-time applications or use in resource-constrained environments. Future work could explore more end-to-end approaches, integrating grounding, ranking, and saliency prediction into a single streamlined model to reduce dependency on individual components and lower computational complexity.

\end{document}